\documentclass{article}
\usepackage[utf8]
{inputenc}

\title{Targeted Learning for Externally Controlled Trials \\and Beyond}
\author{Yixin Fang}
\author{Yixin Fang\thanks{Correspondence: 1 North Waukegan Rd, North Chicago, IL 60064; {email: yixin.fang@abbvie.com}} \quad and \quad Weili He\\
Data and Statistical Sciences, AbbVie Inc.}
\date{\today}

\usepackage{amsmath}
\usepackage{natbib}
\usepackage{graphicx}
\usepackage{amssymb}
\usepackage{setspace}
\usepackage{booktabs}
\usepackage{multirow}
\usepackage{enumitem} 
\usepackage{listings}
\usepackage{comment}
\usepackage{color}
\usepackage{url} 
\newcommand{\indep}{\rotatebox[origin=c]{90}{$\models$}}
\usepackage[a4paper, margin=1.2in]{geometry}

\DeclareMathAlphabet\mathbfcal{OMS}{cmsy}{b}{n}

\begin{document}

\doublespacing

\maketitle

\begin{abstract}

This paper focuses on externally controlled trials (ECTs), which consist of a single-arm clinical trial supplemented by an external control arm (ECA). Potential data sources for the ECA can be broadly classified into three categories: (i) the control arm of a historical clinical trial, (ii) a concurrent cohort study, and (iii) external real-world data (RWD). In this paper, we consider a broad class of estimands defined through balancing weights. For a given estimand in the class, we discuss the efficient influence function (EIF) and the targeted maximum likelihood estimator (TMLE), together with practical implementation algorithms. Within the targeted learning framework, TMLE is a substitution estimator that enjoys several desirable theoretical properties and favorable finite-sample performance. Finally, the proposed framework is extended to two additional settings: (1) randomized controlled trials (RCTs) augmented with external covariate information to define a target population, and (2) indirect treatment comparisons (ITCs) that combine individual patient data (IPD) from an internal RCT with aggregate data (AgD) from an external RCT.

\end{abstract}


\section{Introduction}

The \citet{FDA_ECT_2023} guidance on the design and conduct of externally controlled trials (ECTs) provides recommendations for sponsors and investigators who intend to use ECTs to generate evidence on the safety and effectiveness of drug products. According to the guidance, in an ECT,
\begin{quote}
``(O)utcomes in participants receiving the test treatment according to a protocol are compared to outcomes in a group of people external to the trial who had not received the same treatment. The external control arm can be a group of people, treated or untreated, from an earlier time (historical control), or it can be a group of people, treated or untreated, during the same time period (concurrent control) but in another setting.''
\end{quote}

Thus, an ECT consists of a single-arm clinical trial together with an external control arm (ECA). Consistent with the estimand framework outlined in ICH E9(R1) \citep{ich2020e9}, the study design and the target estimand should be aligned so that the clinical question of interest is clearly defined and can be answered using the collected data. In this section, we review the major sources of ECAs and introduce the class of causal estimands considered throughout this paper.

According to the FDA guidance, ECAs can be broadly classified into three categories:
\begin{enumerate}[label=\roman*.]
    \item A control arm from a historical clinical trial, such as the placebo arm of a conventional randomized controlled trial (RCT), the standard-of-care (SOC) arm of a pragmatic RCT, or the active comparator arm of a head-to-head RCT.
    
    \item A concurrent cohort study conducted under a separate protocol, for example, a cohort of patients receiving SOC in routine clinical practice.

    \item A control arm constructed from external real-world data (RWD), such as electronic health records (EHRs), administrative claims databases, or disease registries, with or without the use of propensity score matching.
\end{enumerate}

Because treatment assignment in an ECT is not randomized, the comparability of the treatment and control populations is critical. As emphasized in the FDA guidance,
\begin{quote}
``Given that externally controlled trials do not involve randomization of the study population to the treatments being compared, the treatment and control arm populations should be as similar as possible regarding known factors that can affect the outcome being measured.''
\end{quote}
The design considerations for achieving this objective differ across the three types of ECAs.

When a historical clinical trial is used to construct the ECA, the design process typically consists of two steps. First, the draft protocol for the single-arm trial is used to identify a historical trial whose eligibility criteria most closely match those of the planned study. Second, after the historical trial has been selected, the protocol for the single-arm trial is refined, whenever feasible, to further align its eligibility criteria with those of the historical trial.

When a concurrent cohort study is used, the cohort study and the single-arm trial should be designed in parallel. In particular, the eligibility criteria, endpoint definitions, baseline covariates, follow-up schedule, and outcome assessment procedures should be harmonized as much as possible to minimize systematic differences between the two study populations.

When RWD are used to construct the ECA \citep{he2023real}, the design process generally begins with a feasibility assessment in which the draft protocol of the single-arm trial is applied to the candidate RWD source. The protocol can then be refined based on the feasibility findings before finalization. In practice, when RWD are obtained from a commercial vendor, budget constraints may limit the amount of data that can be acquired. In such settings, propensity score matching or other matching strategies may be employed to identify and purchase only those external patients who are most comparable to participants in the single-arm trial.

Once the study design has been established, the next step is to specify the target estimand in a manner consistent with the study’s scientific objective. Under the estimand framework described in ICH E9(R1), this process begins with defining the target population. In an ECT, the target population may be the combined study population, in which case the target estimand is the average treatment effect (ATE); the treated population, yielding the average treatment effect on the treated (ATT); or the control population, yielding the average treatment effect on the controls (ATC). The literature on estimating the ATE, ATT, and ATC is extensive \citep{fang2024causal}. 

In this paper, we consider a broad class of estimands for ECTs motivated by the balancing weights proposed by \cite{li2013weighting}. This class includes the ATE, ATT, and ATC as special cases and encompasses several other well-known estimands, including the average treatment effect for the overlap population (ATO)---the population in which the covariate distributions of the treated and control groups overlap---as well as estimands defined for a prespecified target population characterized by a particular covariate distribution. 

Consequently, for each estimand in the proposed class, we develop the corresponding estimators within the targeted learning framework \citep{van2011targeted}. Specifically, we construct an initial estimator of the relevant outcome regression and propensity score, and then apply a targeting step designed to incorporate the information from both treatment groups while directly addressing the estimating equation associated with the target estimand. This framework yields estimators that are tailored to the covariate distribution defining each estimand. 

Regarding the target population, we distinguish between two scenarios. In the first scenario, the distribution of covariates in the target population is estimated from the data, and the resulting estimated target population is treated as fixed when defining the population attribute of the target estimand. For example, if the covariate distribution is estimated using the corresponding empirical distribution, the covariates may be treated as fixed under the conditionality principle \citep{birnbaum1962foundations}. In the second scenario, the estimated target population is viewed as an estimate of an underlying true target population, which is instead regarded as the population attribute of the target estimand. This paper focuses on the first scenario; the second is considered in \cite{fang2026targeted}.

The remainder of the paper is organized as follows. Section~2 introduces the class of balancing-weight estimands that forms the foundation of our framework, following \citet{li2018balancing}. Section~3 develops the corresponding estimators within the targeted learning framework \citep{van2011targeted}. Section~4 extends the proposed methodology to two additional settings. Section~5 presents the implementation algorithms and illustrates their performance through a simulation study. Finally, Section~6 concludes with a discussion.

\section{Estimands} \label{Sec: estimands}

Consider an ECT in which the observed data are represented as
\[
\{(X_i, A_i, Y_i): i=1,\ldots,N=n+m\},
\]
where $X_i$ denotes the vector of baseline covariates, $A_i \in \{0,1\}$ is the treatment indicator ($A_i=1$ for the investigational treatment and $A_i=0$ for the external control), and $Y_i$ denotes the primary outcome. Without loss of generality, we assume that $A_i=1$ for $i=1,\ldots,n$, corresponding to participants enrolled in the single-arm clinical trial, and $A_i=0$ for $i=n+1,\ldots,n+m$, corresponding to participants in the ECA.

\subsection{Identifiablity Assumptions}

Following the double-index notation of \citet{liu2025leveraging}, let
$Y^{a,s}$ denote the potential outcome that would be observed under
treatment assignment $a$ and data source $s$, where $a \in \{0,1\}$ and
$s \in \{0,1\}$. Here, $a=1$ denotes assignment to the investigational
treatment, $a=0$ denotes assignment to the control treatment, $s=1$
denotes participation in the clinical trial, and $s=0$ denotes
participation in the ECA. Thus,
$Y^{1,1}$ and $Y^{0,1}$ represent the potential outcomes under the
investigational and control treatments, respectively, if both treatment
assignments were made within the clinical trial. Similarly,
$Y^{1,0}$ and $Y^{0,0}$ represent the potential outcomes under the
investigational and control treatments, respectively, if the individual
were observed in the external data source.

For an ECT comprising a single-arm clinical trial and an ECA, the
investigational treatment is administered within the clinical trial,
whereas the control outcome is obtained from the external data source.
Thus, the primary estimand of practical interest is often based
on the comparison
\[
Y^{1,1} - Y^{0,0}.
\]
This comparison contrasts the outcome under the investigational treatment
in the clinical-trial population with the outcome under the control
treatment in the external data source. It does not, however, directly
represent the treatment effect that would have been observed if both
treatment arms had been conducted within the clinical trial based on 
\[
Y^{1,1} - Y^{0,1}.
\]
Because the control arm is not observed within the clinical trial, drawing
this latter comparison requires additional assumptions that relate the
unobserved clinical-trial control potential outcome, $Y^{0,1}$, to the
external-source control potential outcome, $Y^{0,0}$. These assumptions
are commonly referred to as transportability or data-source exchangeability
assumptions.

In what follows, we focus on the comparison between the investigational
treatment observed in the clinical trial and the control treatment
observed in the external data source. To simplify notation, we therefore
define
\[
Y^{1} \equiv Y^{1,1}
\qquad\text{and}\qquad
Y^{0} \equiv Y^{0,0}.
\]
Thus, $Y^1$ denotes the potential outcome under the investigational
treatment in the clinical trial, and $Y^0$ denotes the potential outcome
under the control treatment in the external data source. 

We assume that the observed data
\[
\{O_i=(X_i,A_i,Y_i)\}_{i=1}^N
\]
are independent and identically distributed copies of
$O=(X,A,Y)$. The corresponding potential-outcome variables
\[
\{(X_i,A_i,Y_i^1,Y_i^0)\}_{i=1}^N
\]
are assumed to be independent and identically distributed copies of
$(X,A,Y^1,Y^0)$.

Under this notation, we make the following standard assumptions for
identifying the causal contrast between the investigational treatment and
the external control treatment \citep{hernan2020causal}:

\begin{itemize}
    \item Consistency assumption: $Y=AY^{a=1}+(1-A)Y^{a=0}$;
    \item Conditional exchangeability assumption: $(Y^{a=1}, Y^{a=0})\ \indep\ A\mid X$;
    \item Positivity assumption: $\mathbb{P}(A=a\mid X=x)>0$, almost surely under distribution $P_X$.
\end{itemize}

Among these assumptions, conditional exchangeability is particularly challenging to justify in ECTs because treatment assignment is not randomized. As discussed in Section~1, the plausibility of this assumption depends critically on selecting a fit-for-purpose external data source, carefully assessing the comparability between the trial and external populations, and designing the study protocol to maximize their similarity with respect to important prognostic factors.

Define the conditional average treatment effect (CATE) \citep{imbens2015causal} by
\[
\tau(x)=\mathbb{E}(Y^{1}-Y^{0}\mid X=x).
\]
We also define the propensity score \citep{rosenbaum1983central}
\[
g(x)=\mathbb{P}(A=1\mid X=x),
\]
and the outcome regression function
\[
Q(x,a)=\mathbb{E}(Y\mid X=x,A=a).
\]
For notational convenience, we occasionally write $g$ and $Q$ in place of $g(x)$ and $Q(x,a)$, respectively.

\subsection{Target Population}

Our primary interest is the average treatment effect over a prespecified target population \citep{ich2020e9}. Suppose the marginal distribution of $X$ admits a density $f(x)$ with respect to a dominating measure $\mu$. Following \citet{li2018balancing}, we characterize the target population through a nonnegative weighting (or tilting) function $h(x)$, so that the target population has density proportional to $f(x)h(x)$. The corresponding target estimand is
\[
\tau_h=
\frac{\int \tau(x)f(x)h(x)\,\mu(dx)}
{\int f(x)h(x)\,\mu(dx)}.
\]

Different choices of $h(x)$ recover many commonly used causal estimands \citep{li2018balancing}, including
\begin{itemize}
    \item the average treatment effect (ATE), obtained by setting $h(x)=1$;
    \item the average treatment effect on the treated (ATT), obtained by setting $h(x)=g(x)$;
    \item the average treatment effect on the controls (ATC), obtained by setting $h(x)=1-g(x)$;
    \item the average treatment effect for the overlap population (ATO), obtained by setting $
    h(x)=g(x)\{1-g(x)\}$.
\end{itemize}

For the ATE, \(h(x)\) is constant. Consequently, the target population is the combined population, characterized by the empirical distribution of \(X_1,\ldots,X_N\), where
\[
\widehat{f}(X_i)=\frac{1}{N},\qquad i=1,\ldots,N.
\]

For the ATT, if \(g(x)\) is estimated nonparametrically, the target population is the treated population, characterized by the empirical distribution of the treated covariates \(X_1,\ldots,X_n\). If \(g(x)\) is estimated parametrically or semiparametrically, with estimate \(\widehat{g}(x)\), the target population has a density proportional to
\[
\widehat{f}(x)\widehat{g}(x).
\]
Similarly, for the ATC, the corresponding density is proportional to
\[
\widehat{f}(x)\{1-\widehat{g}(x)\}.
\]

For the ATO, if \(g(x)\) is estimated by \(\widehat{g}(x)\), the target population has a density proportional to
\[
\widehat{f}(x)\widehat{g}(x)\{1-\widehat{g}(x)\}.
\]

In the two additional examples in Section 4, \(h(x)\) is not a functional of \(g(x)\); instead, it is also estimated from the covariate distribution.

Thus, for a given application, the corresponding target population is characterized by a density that may be either the empirical covariate distribution or a density proportional to the product of the empirical covariate density and a fixed estimated function \(\widehat{h}(x)\). If \(h(x)\) is prespecified as \(\widehat{h}(x)\), the resulting target estimand is
\[
\left.\tau_h\right|_{h(x)\equiv \widehat{h}(x)}
=
\left.
\frac{\displaystyle\int \tau(x)f(x)h(x)\,\mu(dx)}
{\displaystyle\int f(x)h(x)\,\mu(dx)}
\right|_{h(x)\equiv \widehat{h}(x)}
=
\frac{\displaystyle\int \tau(x)f(x)\widehat{h}(x)\,\mu(dx)}
{\displaystyle\int f(x)\widehat{h}(x)\,\mu(dx)}.
\]

This estimand differs from the estimand considered in \citet{fang2026targeted},
\[
\left.\tau_h\right|_{h(x)\equiv \eta\{g_0(x)\}}
=
\left.
\frac{\displaystyle\int \tau(x)f(x)h(x)\,\mu(dx)}
{\displaystyle\int f(x)h(x)\,\mu(dx)}
\right|_{h(x)\equiv \eta\{g_0(x)\}}
=
\frac{\displaystyle\int \tau(x)f(x)\eta\{g_0(x)\}\,\mu(dx)}
{\displaystyle\int f(x)\eta\{g_0(x)\}\,\mu(dx)},
\]
where \(g_0(x)\) denotes the true, unknown propensity score function.

\subsection{Literature Review}

The estimation of the ATE, ATT, and ATC has been studied extensively in the causal inference literature \citep{hernan2020causal,van2011targeted,fang2024causal}. Balancing-weight estimands, including the ATO, have also attracted considerable attention \citep{li2018balancing}. Additional estimands have been proposed to address practical challenges such as limited covariate overlap. For example, \citet{liu2024average} introduced the overlap-weighted average treatment effect on the treated (OWATT). Furthermore, alternative balancing weight functions, including entropy weights \citep{zhou2020propensity} and matching weights \citep{liliang2013weighting}, have been proposed to improve covariate balance and estimation efficiency under different settings.

Since its introduction by \citet{li2018balancing}, the weighting function $h(x)$ has become widely known as a \emph{tilting function}. The tilting-function framework has subsequently been generalized and applied in numerous settings. For example, \citet{yang2021subgroup} extended the framework to causal subgroup analysis, \citet{liu2025positivity} provided a comprehensive tutorial on propensity score weighting under limited positivity, and \citet{zhou2022psweight} developed the \texttt{PSweight} \textsf{R} package to implement a broad class of balancing-weight analyses. These developments demonstrate the versatility of the tilting-function framework and motivate the development of targeted learning methods for this general class of estimands.

Given this substantial body of work, the objective of this paper is not to review the existing estimation methods for these individual estimands. Instead, we develop targeted maximum likelihood estimators (TMLEs) for this general class of balancing-weight estimands, thereby providing a unified estimation framework for each individual estimand. Within the targeted learning framework, TMLE is a substitution estimator that enjoys desirable theoretical properties, while also exhibiting favorable finite-sample performance \citep{van2011targeted}.

Although our presentation focuses on externally controlled trials, the proposed framework is considerably more general. The estimands, efficient influence functions, and TMLEs developed in this paper apply equally to observational studies and other nonrandomized settings, provided that the required causal identification assumptions are appropriate for the scientific question and the relevant sources of confounding have been adequately addressed.

\section{Estimators} \label{Sec: estimators}

Recall that the target estimand is defined as $\tau_h$, where $h(x)$ is either a known function or is fixed at its empirical estimate. In practice, the function $h(x)$ is prespecified by the user, who may choose a known function, such as $h(x)\equiv 1$ for the estimation of ATE, or specify $h(x)$ as a functional of the estimated propensity score, such as $h(x) \equiv \widehat{g}(x)$
for the estimation of ATT, $h(x) \equiv \{1-\widehat{g}(x)\}$
for the estimation of ATC, and $h(x) \equiv \widehat{g}(x)\{1-\widehat{g}(x)\}$
for the estimation of ATO.

\subsection{Identification}

We consider two complementary strategies for identifying $\tau_h$ \citep{fang2020two}: standardization and weighting. By the standardization, strategy, the target estimand can be written as
\begin{align*}
    \tau_h
    =
    \mathbb{E}\left\{ h^*(X)\big[ Q(X,1)-Q(X,0)\big]\right\},
\end{align*}
where the normalized tilting function is defined by
\begin{align*}
    h^*(x)
    =
    \frac{h(x)}{\int f(x)h(x)\,\mu(dx)}.
\end{align*}
This representation expresses $\tau_h$ as a weighted average of the conditional treatment effect.

Alternatively, by the weighting strategy, $\tau_h$ can be expressed as
\begin{align*}
    \tau_h
    =
    \mathbb{E}\left[
    \frac{h^*(X)AY}{g(X)}
    -
    \frac{h^*(X)(1-A)Y}{1-g(X)}
    \right].
\end{align*}
This form highlights the role of inverse probability weighting under the target distribution.

The standardization representation motivates maximum likelihood plug-in estimator (MLE; a.k.a.~the g-computation estimator)
\begin{align*} 
\widehat{\tau}_h^{\mathrm{MLE}}
=
\frac{1}{\sum_{i=1}^N h(X_i)}
\sum_{i=1}^N h(X_i)\Big[
\widehat{Q}(X_i,1)-\widehat{Q}(X_i,0)
\Big],
\end{align*}
which is consistent provided that the outcome regression estimator $\widehat{Q}(x,a)$ is consistent, and is therefore singly robust.

Similarly, the weighting representation yields the inverse probability weighting (IPW) estimator
\begin{align*} 
\widehat{\tau}_h^{\mathrm{IPW}}
=
\frac{1}{\sum_{i=1}^N h(X_i)}
\sum_{i=1}^N
\left[
\frac{h(X_i)A_iY_i}{\widehat{g}(X_i)}
-
\frac{h(X_i)(1-A_i)Y_i}{1-\widehat{g}(X_i)}
\right],
\end{align*}
which is consistent if the propensity score estimator $\widehat{g}(x)$ is consistent, and is also singly robust.

\subsection{Efficient Influence Function}

The efficient influence function (EIF) is central to the construction of semiparametrically efficient estimators \citep{bickel1993efficient, tsiatis2006semiparametric, hernan2020causal, fang2024causal}. As shown in Appendix~A.1, the EIF for $\tau_h$ is given by
\begin{align*}
\phi_h(X,A,Y)
&=
h^*(X)\big[Q(X,1)-Q(X,0)-\tau_h\big] \\
&\quad +
\left[
\frac{h^*(X)A}{g(X)}
-
\frac{h^*(X)(1-A)}{1-g(X)}
\right]\big[Y-Q(X,A)\big].
\end{align*}

This representation decomposes the EIF into a component driven by the conditional mean model and a mean-zero correction term that accounts for estimation of the nuisance functions. The same influence function has appeared in several related settings; see, for example, \citet{hirano2003efficient}, \citet{wang2025rate}, and \citet{liu2025leveraging}.

\subsection{Targeted Learning}

After we derive the EIF, we now follow the targeted learning framework \citep{van2011targeted} to develop the targeted maximum likelihood estimator (TMLE) for $\tau_h$. The framework consists of two main steps: the initial step and the target step. 

\subsubsection{Binary Outcome}

In the initial step, we fit a generalized linear model or a super learner \citep{van2007super} for the outcome regression function $Q(x, a)$ and the propensity score function $g(x)$, respectively, leading to initial estimates $\widehat{Q}^{0}(x, a)$ and $\widehat{g}^{0}(x)$. To fit a super learner, it is crucial to specify a library of base models follow the flowchart proposed by \cite{phillips2023practical}. Thus, we obtain the following initiate estimator of the estimand, 
\begin{align*}
    \widehat{\tau}_h^0=\frac{1}{\sum_{j=1}^N h(X_j)}\sum_{i=1}^N {\color{black} h(X_i)}\left[\widehat{Q}^0(X_i, 1)-\widehat{Q}^0(X_i, 0)\right].
\end{align*}

In the target step, according to the efficient influence function $\phi_h(X, A, Y)$, the following least favorable parametric submodel is considered: 
\begin{gather*}
    f(X;\epsilon_1)=\widehat{f}^0(X)\left[1+\epsilon_1 \cdot D(X)\right],\\
    g(X; \epsilon_2)]=\mbox{expit}\left\{\mbox{logit}[\widehat{g}^0(X)]+\epsilon_2\cdot 0\right\},\\
    Q(X, A;\epsilon_3)= \mbox{expit}\left\{\mbox{logit}[\widehat{Q}^0(X, A)]+\epsilon_3\cdot H(X, A)\right\}, 
\end{gather*}
where $\mbox{expit}(x)=\frac{e^{x}}{1+e^{x}}$, $\mbox{logit}(x)=\log(\frac{x}{1-x})$, $\widehat{f}^0$ is the empirical distribution of $\{X_1, \dots, X_N\}$, $D(X)=h(X)[\widehat{Q}^0(X, 1)-\widehat{Q}^0(X, 0)-\widehat{\tau}_h^0]$, and the clever covariate is defined as
\begin{align*}
    H(X, A)=\frac{h(X)A}{\widehat{g}^0(X)}-\frac{h(X)(1-A)}{1-\widehat{g}^0(X)}.
\end{align*}
Note that the normalization constant for $h(x)$ is excluded from the definition of the clever covariate because it can be absorbed into $\epsilon_3$. 

Since $\widehat{f}^0$ is nonparametric maximum likelihood estimator (NPMLE) and the score of $g(X; \epsilon_2)$ at zero is zero, the initiate estimates of {\color{black} $\widehat{f}^0$} and $\widehat{g}^0$ are not updated. Only the regression function estimate $\widehat{Q}(X, A)$ will be updated via the following logistic regression:
\begin{align*}
    \mbox{logit}[Q(X, A; \epsilon_3)]=\mbox{logit}[\widehat{Q}^0(X, A)]+\epsilon_3\cdot H(X, A).
\end{align*}
Let $\widehat{\epsilon}_3$ be the resulting estimate of $\epsilon_3$, leading to the updated estimate of $Q(X, A)$, 
\begin{align*}
    \widehat{Q}^*(X, A)=\mbox{expit}\left\{\mbox{logit}[\widehat{Q}^0(X, A)]+\widehat{\epsilon}_3\cdot H(X, A)\right\}.
\end{align*}
From the estimating equation of the above logistic regression, only one round of updating is needed because the maximum likelihood estimate of $\epsilon_3$ in the next round would be zero \citep{van2011targeted}. Thus, after the target step, we obtain the TMLE: 
\begin{align*}
    \widehat{\tau}_h=\frac{1}{\sum_{j=1}^N h(X_j)}\sum_{i=1}^Nh(X_i)\left[\widehat{Q}^*(X_i, 1)-\widehat{Q}^*(X_i, 0)\right].
\end{align*}

\subsubsection{Bounded Continuous Outcome}

The above TMLE procedure can be extended for bounded continuous outcome $Y\in [a, b]$. First, consider the following transform:
\begin{align*}
    \widetilde{Y}=\frac{Y-a}{b-a}.
\end{align*}
Since $\widetilde{Y}\in[0,1]$, we can use the above procedure to obtain an TMLE, denoted by $\widetilde{\tau}_h$. Thus, using the inverse of the linear transform, we obtain the TMLE for the target estimand $\tau_h$, 
\begin{align*}
    \widehat{\tau}_h=(b-a)\widetilde{\tau}_h.
\end{align*}

Note that the above procedure can be applied to ordinal outcome. 

\subsubsection{Time-to-Event Outcome}

For a time-to-event outcome, the estimand of interest may take several forms, including the hazard ratio \citep{hernan2010hazards}, treatment-specific survival curves \citep{rytgaard2024one}, the survival-rate difference or ratio \citep{van2011targeted}, and the restricted mean survival time (RMST) difference or ratio \citep{karrison1987restricted}. Recently, overlap-weighted survival methods have been proposed; see, for example, \cite{cheng2022addressing} and \cite{cao2025using}.

A detailed discussion of survival analysis is beyond the scope of this paper and will be presented elsewhere. Here, we briefly note that the proposed TMLE procedure is applicable when the estimand of interest is the RMST difference on the target population, possibly tilted by a prespecified weighting function $h(x)$. The key idea is to leverage the so-called pseudo-observations \citep{andersen2004regression, zeng2023pseudo, cao2024overlap}. If we denote the pseudo-observations by $\widetilde{Y}$, then the TMLE procedure above extends straightforwardly by replacing $Y$ with $\widetilde{Y}$. See \cite{jin2026framework} for further details.

\subsubsection{Standard Error} \label{Sec:SE}

According to the asymptotic normality of TMLE \citep{van2011targeted}, we can consider the following as the estimator for the standard error of $\widehat{\tau}_h$,
\begin{align*}
\frac{1}{\sqrt{N}}\sqrt{\frac{1}{N}\sum_{i=1}^N\widehat{\phi}^2_h(X_i, A_i, Y_i)},   
\end{align*}
where $\widehat{\phi}_h(X_i, A_i, Y_i)$ is the estimated efficient influence function, 
\begin{align*}
    \widehat{h}^*(X_i)\left[\widehat{Q}^*(X_i, 1)-\widehat{Q}(X_i, 0)-\widehat{\tau}_h\right]+\left[\frac{\widehat{h}^*(X_i)A_i}{\widehat{g}^0(X_i)}-\frac{\widehat{h}^*(X_i)(1-A_i)}{1-\widehat{g}^0(X_i)}\right][Y_i-\widehat{Q}^*(X_i, A_i)],
\end{align*}
where \begin{align*}
    \widehat{h}^*(X_i)=\frac{h(X_i)}{\frac{1}{N}\sum_{j=1}^N h(X_j)}.
\end{align*}

\subsection{Remarks}

\paragraph{Remark 1: We can consider risk ratio and odds ratio estimands for binary outcomes.} 
In addition to the risk difference, the framework can naturally accommodate other causal estimands for binary outcomes, such as the risk ratio and the odds ratio. Once the targeted estimate $\widehat{Q}^*(X,A)$ has been obtained, one may define the TMLE of the risk ratio as
\begin{align*}
    \widehat{\mathrm{RR}}
    =
    \frac{\sum_{i=1}^N \widehat{h}^*(X_i)\widehat{Q}^*(X_i,1)}
         {\sum_{i=1}^N \widehat{h}^*(X_i)\widehat{Q}^*(X_i,0)}.
\end{align*}
Similarly, the TMLE of the odds ratio can be constructed from the corresponding estimated risks,
\begin{align*}
    \widehat{\mathrm{OR}}
    =
    \frac{
        \left\{\sum_{i=1}^N \widehat{h}^*(X_i)\widehat{Q}^*(X_i,1)\right\}
        \left\{1-\sum_{i=1}^N \widehat{h}^*(X_i)\widehat{Q}^*(X_i,0)\right\}
    }{
        \left\{1-\sum_{i=1}^N \widehat{h}^*(X_i)\widehat{Q}^*(X_i,1)\right\}
        \left\{\sum_{i=1}^N \widehat{h}^*(X_i)\widehat{Q}^*(X_i,0)\right\}
    }.
\end{align*}
The standard errors of these estimators may be obtained using the Delta method \citep{van2000asymptotic}, or alternatively through the empirical influence function when available.

\paragraph{Remark 2: Missing data handling.}

Assume that the outcome variable may be subject to missingness. Let $\Delta$ denote the indicator of non-missingness, where $\Delta=1$ if $Y$ is observed and $\Delta=0$ otherwise. Suppose that $(X_i, A_i, Y_i^{a=1}, Y_i^{a=0})$ are i.i.d.\ copies of $(X, A, Y^{a=1}, Y^{a=0})$, and that $(X_i, A_i, \Delta_i, \Delta_i Y_i)$ are i.i.d.\ copies of $(X, A, \Delta, \Delta Y)$.

To handle missing data, we assume the following assumptions: 
\begin{itemize}
    \item Consistency assumption: If $\Delta=1$, $Y=AY^{a=1}+(1-A)Y^{a=0}$;
    \item Conditional exchangeability assumption: $(Y^{a=1}, Y^{a=0})\ \indep (A, \Delta)\mid X$;
    \item Positivity assumption: $\mathbb{P}(A=a, \Delta=1\mid X=x)>0$, almost surely under $P_X$.
\end{itemize}

To build intuition, we may view $\Delta=1$ as an \emph{intervention} enforcing observation of the outcome, and $(A=a,\Delta=1)$ as a joint intervention assigning treatment $a$ and ensuring observation \citep{fang2026every}. Under this interpretation, the above assumptions naturally extend the standard identification conditions in Section~2. In particular, conditional exchangeability implies a missing-at-random (MAR) mechanism.

Under these assumptions, the estimand $\tau_h$ remains unchanged and admits the following identification:
\begin{align*}
    \tau_h
    &= \mathbb{E}\!\left[h^*(X)\{Y^{a=1}-Y^{a=0}\}\right] \\
    &= \mathbb{E}\!\left\{\mathbb{E}\!\left[h^*(X)\{Y^{a=1}-Y^{a=0}\}\mid X\right]\right\} \\
    &= \mathbb{E}\!\left\{h^*(X)\Big(\mathbb{E}[Y^{a=1}\mid X, A=1, \Delta=1]
        - \mathbb{E}[Y^{a=0}\mid X, A=0, \Delta=1]\Big)\right\} \\
    &= \mathbb{E}\!\left\{h^*(X)\Big(\mathbb{E}[Y\mid X, A=1, \Delta=1]
        - \mathbb{E}[Y\mid X, A=0, \Delta=1]\Big)\right\}.
\end{align*}

Define the missingness propensity score and outcome regression under observed outcomes as
\begin{align*}
    g_{\delta}(x,a) &= \mathbb{P}(\Delta=1 \mid X=x, A=a), \\
    Q_{\Delta=1}(x,a) &= \mathbb{E}(Y \mid X=x, A=a, \Delta=1).
\end{align*}

Similarly, the efficient influence function is given by
\begin{align*}
\phi_h(X,A,\Delta,\Delta Y)
&= h^*(X)\left[Q_{\Delta=1}(X,1)-Q_{\Delta=1}(X,0)-\tau_h\right] \\
&\quad + \left[\frac{h^*(X)A\Delta}{g(X)\,g_{\delta}(X,1)}
- \frac{h^*(X)(1-A)\Delta}{(1-g(X))\,g_{\delta}(X,0)}\right]
\left[Y - Q_{\Delta=1}(X,A)\right].
\end{align*}

For intuition, note that we may again interpret $(A=a,\Delta=1)$ as a joint intervention assigning treatment $a$ and ensuring observation \citep{fang2026every}. By the chain rule of probability,
\[
\mathbb{P}(A=a,\Delta=1\mid X)
=
\mathbb{P}(A=a\mid X)\,\mathbb{P}(\Delta=1\mid A=a,X).
\]

The TMLE can be constructed analogously to Subsection~3.3, with the additional step of estimating $g_{\delta}$ via Super Learner to obtain $\widehat{g}^0_{\delta}$. The corresponding clever covariate is
\begin{align*}
H(X,A,\Delta)
=
\frac{h(X)A\Delta}{\widehat{g}^0(X)\widehat{g}^0_{\delta}(X,1)}
-
\frac{h(X)(1-A)\Delta}{\{1-\widehat{g}^0(X)\}\widehat{g}^0_{\delta}(X,0)}.
\end{align*}

\section{Extensions}

The proposed targeted learning framework for estimating $\tau_h$ has thus far been described for ECTs. It can be naturally extended to the following two settings. In both settings, the function $h$ is prespecified by the user and treated as fixed. The corresponding target estimand is defined with respect to a target population whose density is proportional to the product of the marginal density $f(x)$ and the prespecified function $h(x)$.

\subsection{RCT Augmented with External Covariates}

Assume there are data $\{(X_i, A_i, Y_i), i=1, \dots, n\}$ from an RCT, where $A_i$ is the treatment assignment variable (1 for investigational treatment and 0 for placebo; via randomization or stratified randomization). Since RCTs are often designed with strict inclusion and exclusion criteria, a disadvantage is the lack of generalizability. To account for the lack of generalizability, we can combine an RCT with an observational study. As reviewed in \cite{colnet2024causal}, there are three types of observational studies can be considered 
\begin{enumerate}
    \item The observational study only has covariates data;
    \item The observational study only has control data; 
    \item The observational study has both control data and treated data. 
\end{enumerate}

Here we focus on the first type: an RCT augmented with external covariates. As reviewed by \cite{colnet2024causal}, several methods are available for this study design, including inverse probability of sampling weighting (IPSW) \citep{cole2010generalizing}, the plug-in g-formula \citep{dahabreh2021study}, calibration weighting (CW) \citep{lee2023improving}, augmented IPSW (AIPSW) \citep{colnet2022causal}, and augmented CW (ACW) \citep{lee2023improving}. Notably, AIPSW and ACW offer double robustness. However, there is limited literature on applying the targeted learning procedure in this context. This paper aims to fill that gap. 

Let $\mathcal{R}=\{(X_i, A_i, Y_i, S_i=1), i=1, \dots, n\}$ be the data from the RCT and $\mathcal{O}=\{(X_i, S_i=0), i=1, \dots, m\}$ be the data from the observational study. 

Define the following probability of sampling: 
\begin{align*}
    s(x)=\mathbb{P}(S=1\mid X=x).
\end{align*}

Using a logistic regression model or a super learner, an estimate of the propensity score $s(x)$ is obtained, defined as $\widehat{s}^0(x)$. Thus, define 
\begin{align*}
    h(x)=\frac{1-\widehat{s}^0(x)}{\widehat{s}^0(x)}.
\end{align*}

Furthermore, the estimand of interest is also defined as $\tau_h$, 
\begin{align*}
    \tau_h=\frac{\mathbb{E}_{X\mid S=1}\{h(X)[Y^{a=1, s=1}-Y^{a=0, s=1}]\}}{\mathbb{E}_{X\mid S=1}[h(X)]}.
\end{align*}
Consequently, the proposed TMLE procedure can be applied straightforwardly. 

By the definition of the above $h(x)$, the target population behind $\tau_h$ is the one whose distribution is proportional to 
\begin{align*}
    \mathbb{P}(X=x\mid S=1)h(x)&=\mathbb{P}(X=x\mid S=1)\times\frac{\mathbb{P}(S=0\mid X=x)}{\mathbb{P}(S=1\mid X=x)}\\
     &=\mathbb{P}(X=x\mid S=1)\times\frac{\mathbb{P}(S=0, X=x)}{\mathbb{P}(S=1, X=x)}\\
    &=\mathbb{P}(X=x\mid S=1)\times\frac{\mathbb{P}(X=x\mid S=0)}{\mathbb{P}(X=x\mid S=1)}\times\frac{\mathbb{P}(S=0)}{\mathbb{P}(S=1)} \\
    &=\mathbb{P}(X=x\mid S=0)\times\frac{\mathbb{P}(S=0)}{\mathbb{P}(S=1)} \\
    &\propto \mathbb{P}(X=x\mid S=0).
\end{align*}
Therefore, $\tau_h$ measures the in-trial treatment effect averaged over the observational population.

\subsection{Indirect Treatment Comparison}

In support of Health Technology Assessment (HTA) submission, we often need to conduct indirect treatment comparisons (ITC). One common type of ITC is population-adjusted indirect comparisons (PATC), in which individual patient data (IPD) in one trial and aggregate data (AgD) in the other trial are used to adjust for differences between trials in the distribution of variables that influence outcome \citep{phillippo2016nice}. The most popular PATC method is matching adjusted indirect comparison (MAIC) \citep{signorovitch2010comparative, phillippo2016nice}.

Assume there is an internal RCT generating IPD, $\{(X_i, A_i, Y_i), i=1, \dots, n\}$, where $A_i$ is the treatment assignment variable (1 for investigational treatment and 0 for placebo). In addition, there is an external RCT with only AgD that compares two treatments: the external investigational treatment denoted as 2 and the external placebo denoted as 0. The AgD includes the average covariates $\overline{X}_E$, the point estimate of the ATE comparing 2 with 0 denoted as $\widehat{\tau}_E^{20}$ along with an estimate of the standard error $\widehat{\sigma}_E^{20}$. Here the subscript `E' emphasizes that the target population is the one underlying the external RCT. 

Let $S=1$ and $0$ indicate the internal RCT and external RCT, respectively. We attempt to define an $h(x)$ such that the weighted distribution of $X$ in the internal RCT is the same as that in the external RCT, that is, 
\begin{align*}
    \mathbb{P}(X=x\mid S=1)h(x)=\mathbb{P}(X=x\mid S=0).
\end{align*}
Since we only have $\overline{X}_E$ from the external RCT, we attempt to define an $h(x)$ such that 
\begin{align*}
    \int x\mathbb{P}(X=x\mid S=1)h(x)\mu(dx)=\int x\mathbb{P}(X=x\mid S=0)\mu(dx).
\end{align*}
This comes to the idea of MAIC \citep{signorovitch2010comparative}, which proposes to define an $h(x)=\exp(X^\top\beta)$ such that 
\begin{align*}
    \frac{1}{n}\sum_{i=1}^nX_ih(X_i)=\frac{1}{n}\sum_{i=1}^nX_i\exp(X_i^\top\beta)=\overline{X}_E.
\end{align*}
Define the solution to the above equation as $\widehat{\beta}$, resulting in the following chosen $h(x)$, 
\begin{align*}
    h_E(x)=\exp(x^\top\widehat{\beta}).
\end{align*}

Therefore, for the given $h_E(x)$, the estimand of interest is defined as $\tau^{10}_{h_E}$, which measures the ATE comparing 1 with 0 averaged over the external population, that is, the one underlying the external RCT. Consequently, the proposed TMLE procedure can be applied straightforwardly to produce an estimate $\widehat{\tau}^{10}_{h_E}$ along with an estimate of the standard error $\widehat{\sigma}^{10}_{h_E}$.

Combing the two results, we obtain the point estimate for the ATE comparing treatment 1 with treatment 2 averaged over the external population:, 
\begin{align*}
    \widehat{\tau}^{12}_E=\widehat{\tau}^{10}_{h_E}-\widehat{\tau}^{20}_E,
\end{align*}
along with an estimate of its standard error, 
\begin{align*}  \widehat{\sigma}_E^{12}=\sqrt{\left(\widehat{\sigma}^{10}_{h_E}\right)^2+\left(\widehat{\sigma}^{20}_E\right)^2}.
\end{align*}

Please refer to \cite{phillippo2016nice} for a more detailed review of MAIC and the choice of
$h(x)=\exp\{x^\top\beta\}$. Because only aggregate data are available from the external trial,
the simplest form of $h(x)$ is more appropriate than more complex alternatives.

\section{Numerical Results}

\subsection{Implementation}

We now outline the implementation. For simplicity, we consider a binary outcome without missing outcome data. The approach generalizes to bounded continuous outcomes through an appropriate linear transformation, to time-to-event outcomes through pseudo-observations, and to incomplete data settings by augmenting the procedure with a model for the non-missingness probability.

To address missingness in baseline covariates, we adopt the indicator method described in \cite{fang2024causal}. For each covariate with missing values, continuous variables are imputed by their sample mean and binary variables by their majority category. A missingness indicator is created for each such covariate to denote whether the original value was missing. With indicators as additional covariates, let $W$ denote the resulting augmented covariate vector. The functions originally defined in terms of $X$ are then re-expressed as functions of $W$, namely $h(W)$, $Q(W, A)$, and $g(W)$. This notation is consistent with that adopted in \cite{gruber2012tmle}.

For each specified function $h$, the TMLE procedure for estimating $\tau_h$ proceeds in two steps. In the initial step, we estimate the outcome regression $Q(W, A)$ and the propensity score $g(W)$ using either a generalized linear model (GLM) or super learner (SL). Upon completion of this step, the observed data are augmented with the corresponding fitted values, resulting in the dataset summarized in Table~\ref{Tab:data-initial}.

\begin{table}[ht]\caption{The tentative dataset from the initial step}\label{Tab:data-initial}
\medskip
\centering
\renewcommand{\arraystretch}{1}
\begin{tabular}{c cccccc ccccc }
\toprule
ID & $\texttt{W}$ & $\texttt{A}$ & $\texttt{Y}$ & $\texttt{QAW}$ & $\texttt{Q1W}$ & $\texttt{Q0W}$ & $\texttt{g1W}$ & $\texttt{g0W}$  \\
\midrule
1 & $W_1$ & $A_1$ & $Y_1$ & $\widehat{Q}^0(W_1, A_1)$ & $\widehat{Q}^0(W_1, 1)$ & $\widehat{Q}^0(W_1, 0)$ & $\widehat{g}^0(W_1)$ & $1-\widehat{g}^0(W_1)$ \\
2 & $W_2$ & $A_2$ & $Y_2$ & $\widehat{Q}^0(W_2, A_2)$ & $\widehat{Q}^0(W_2, 1)$ & $\widehat{Q}^0(W_2, 0)$ & $\widehat{g}^0(W_2)$ & $1-\widehat{g}^0(W_2)$ \\
$\cdots$ &\\
$N$ & $W_N$ & $A_N$ & $Y_N$ & $\widehat{Q}^0(W_N, A_N)$ & $\widehat{Q}^0(W_N, 1)$ & $\widehat{Q}^0(W_N, 0)$ & $\widehat{g}^0(W_N)$ & $1-\widehat{g}^0(W_N)$ \\
\bottomrule
\end{tabular}
\end{table}

In the target step, we define the clever covariate, fit a logistic regression model, obtain an estimate of $\epsilon_3$, and update the regression function estimates $\widehat{Q}^0$ to $\widehat{Q}^*$. The partial \textsf{R} code for implementing this step is provided in Appendix A.2. 
After completing this step, additional columns that are presented in Table~\ref{Tab:data-clever} can be appended to Table~\ref{Tab:data-initial}. The TMLE estimate can be obtained from the first column and the last two columns of Table~\ref{Tab:data-clever}:
\begin{align*}
    \widehat{\tau}_h=\frac{1}{\sum_{j=1}^N h(W_j)}\sum_{i=1}^Nh(W_i)\left[\widehat{Q}^*(W_i, 1)-\widehat{Q}^*(W_i, 0)\right].
\end{align*}

\begin{table}[ht]\caption{The tentative dataset from the target step}\label{Tab:data-clever}
\medskip
\centering
\renewcommand{\arraystretch}{1}
\begin{tabular}{cccccc }
\toprule
ID & $\texttt{h}$ & $\texttt{HAW}$ & $\texttt{logitQAW}$ & $\texttt{Qstar1W}$ & $\texttt{Qstar0W}$   \\
\midrule
1 & $h(W_1)$ & $H(W_1, A_1)$ & $\mbox{logit}[\widehat{Q}^0(W_1, A_1)]$ & $\widehat{Q}^*(W_1, 1)$ & $\widehat{Q}^*(W_i, 0)$ \\
2 & $h(W_2)$ & $H(W_2, A_2)$ & $\mbox{logit}[\widehat{Q}^0(W_2, A_2)]$ & $\widehat{Q}^*(W_2, 2)$ & $\widehat{Q}^*(W_2, 0)$ \\
$\cdots$ &\\
N & $h(W_N)$ & $H(W_N, A_N)$ & $\mbox{logit}[\widehat{Q}^0(W_N, A_N)]$ & $\widehat{Q}^*(W_N, N)$ & $\widehat{Q}^*(W_N, 0)$ \\
\bottomrule
\end{tabular}
\end{table}

\subsection{Simulation} \label{sec:simulation}

We conducted simulation studies to investigate the performance of the proposed TMLE for a given $h(x)$ under four nuisance-model scenarios: 
\begin{itemize}
\item[(i)] the outcome regression $Q$ was misspecified and the propensity score $g$ was correctly specified; 
\item[(ii)] $Q$ was correctly specified and $g$ was misspecified; 
\item[(iii)] both $Q$ and $g$ were misspecified; 
\item[(iv)] both $Q$ and $g$ were correctly specified.
\end{itemize}

In this section, we present simulation experiments for a ECT data setting with poor covariate overlap between the treated and external-control population. 

\subsubsection{Data Generating Mechanism} \label{sec:datagen}

 Each replication contained $n_1=200$ treated trial participants and $n_0=300$ external controls. For treated participants ($A=1, S=1$), the covariate $X = (X_1,X_2,X_3,X_4)$, where $X_j$'s were independently generated by 
\begin{align*}
X_1 \sim \operatorname{Bernoulli}(0.68),\quad
X_2 \sim N(0.75,1.00^2),\quad
X_3 \sim N(0.45,0.80^2),\quad
X_4 \sim \operatorname{Uniform}(-1,1),
\end{align*}
whereas for external controls ($A=0, S=0$),
\begin{align*}
X_1 \sim \operatorname{Bernoulli}(0.32),\quad
X_2 \sim N(-0.55,0.85^2),\quad
X_3 \sim N(-0.35,1.15^2),\quad
X_4 \sim \operatorname{Uniform}(-1,1).
\end{align*}

The binary outcome was generated from
\[
Y\mid X,A\sim\operatorname{Bernoulli}\{Q_0(X,A)\},
\]
where
\begin{align*}
\operatorname{logit}\{Q_0(X,0)\}
={}&-0.80+0.40X_1+1.30\sin(1.50X_2)-0.25X_3\\
&+0.75(X_3^2-1)+0.55X_2X_4,
\end{align*}
and
\begin{align*}
\operatorname{logit}\{Q_0(X,1)\}
={}&\operatorname{logit}\{Q_0(X,0)\}
+0.40+0.40X_1\\
&+1.10\sin(1.80X_2)-0.80(X_3^2-1)+0.55X_2X_4.
\end{align*}

Assuming arm membership is randomly assigned with $\mathbb{P}(A=1) = \frac{n_1}{n_1 + n_0}$, we can apply Bayes' rule to compute the closed form of the propensity score. Because $X_4$ follows the same $\text{Uniform}(-1,1)$ distribution in both arms, it carries no information on the arm membership and is dropped out of the likelihood ratio between the two source densities. Consequently, the true propensity score $g(x)=\mathbb{P}(A=1\mid X=x)$ takes the closed form
\begin{equation*} 
   \mathbb{P}(A=1 \mid X) = g(X) = \text{expit}\{-1.14 + 1.51X_1 + 1.51X_2 + 0.19X_2^2 + 0.97X_3 - 0.40X_3^2\}.
\end{equation*}

\subsubsection{Target Estimand}

The prespecified tilting function was
\[
h(x)=\exp\{-0.18x_2^2-0.10x_3^2\}.
\]
For the design mixture \(f(x)=0.4f_1(x)+0.6f_0(x)\), the target estimand was
\begin{equation*}
\tau_h=
\frac{\mathbb{E}\left[h(X)\{Q_0(X,1)-Q_0(X,0)\}\right]}
{\mathbb{E}\{h(X)\}}.
\end{equation*}
The common truth was approximated using 5,000,000 independent covariate draws, including 2,000,000 treated population draws and 3,000,000 external population draws:

\[
\tau_h^{\mathrm{MC}}=0.136.
\]

\subsubsection{Estimators}

The proposed TMLE was compared with $\widehat{\tau}_h^{\mathrm{MLE}}$ and $\widehat{\tau}_h^{\mathrm{IPW}}$ in Section 3.1, and the following augmented IPW (AIPW) estimator,
\begin{align*}
\widehat\tau_h^{\mathrm{AIPW}}
={}&\frac{1}{N}\sum_{i=1}^N\widehat h_i^*(X_i)
\Bigg[
\widehat Q(X_i,1)-\widehat Q(X_i,0)\nonumber\\
&+
\frac{A_i}{\widehat g(X_i)}
\{Y_i-\widehat Q(X_i,1)\}
-
\frac{1-A_i}{1-\widehat g(X_i)}
\{Y_i-\widehat Q(X_i,0)\}
\Bigg].
\end{align*}

\subsubsection{Working models} \label{Sec: working models}

The correctly specified outcome model contained $A$, $X_1$, $\sin(1.40X_2)$, $\tanh(X_3)$, $X_3^2$, $X_2X_4$, and the treatment interactions $AX_1$, $A\sin(1.70X_2)$, $AX_3^2$, and $AX_2X_4$. The misspecified outcome model retained only $A$, $X_1$, $X_2$, $X_3$, and $X_4$.

The correctly specified log density-ratio model for $g(X)$ contained $X_1$, $X_2$, $X_3$, $X_2^2$, and $X_3^2$. The misspecified propensity score model retained only linear main effects of $X_1$, $X_2$, $X_3$, and $X_4$.

All estimated propensity scores were truncated to $[0.025,0.975]$ to follow the software default of the \textsf{R} package ``tmle" \citep{gruber2012tmle}. Therefore, ``correctly specified $g$'' refers to the parametric model before truncation.

\subsubsection{Performance measures and inference} \label{sec:inference}

For $B=1{,}000$ replications, absolute bias was
\[
\text{Abs.~bias} = \left|\frac{1}{B}\sum_{b=1}^B\widehat\tau_h^{(b)}
-\tau_h^{\mathrm{MC} (b)}\right|,
\]
and root mean squared error (RMSE) was
\[
\text{RMSE} = \sqrt{
\frac{1}{B}\sum_{b=1}^B
\left(\widehat\tau_h^{(b)}-\tau_h^{\mathrm{MC} (b)}\right)^2
}.
\]

For TMLE inference, we evaluated the EIF-Wald confidence interval and the target bootstrap (TB) intervals \citep{van2018targeted}. In the target bootstrap methods, the initial nuisance estimates were held fixed, and only the clever logistic regression was refitted in each of 300 stratified bootstrap samples. The ``TB-Wald" estimated the SE and then used it to form a Wald test to construct the confidence interval. The ``TB-percentile" used the empirical percentiles to construct the confidence intervals.

\subsubsection{Results}

The results for the data setting in Section \ref{sec:datagen} are summarized in Table \ref{tab:four-scenario-point} and Table \ref{tab:four-scenario-inference}. 

\begin{table}[htbp]
\centering
\caption{Point-estimation performance for a given $h$}
\label{tab:four-scenario-point}
\begin{tabular}{llrrrrr}
\toprule
Scenario & Method & Truth & Mean & Abs.~bias & EmpSD & RMSE\\
\midrule
\multirow{4}{*}{$Q$ misspecified; $g$ correct$^*$} & MLE  & 0.1363 & 0.2363 & 0.1000 & 0.0609 & 0.1171\\
 & IPW  & 0.1363 & 0.1350 & 0.0012 & 0.0593 & 0.0593\\
 & AIPW & 0.1363 & 0.1495 & 0.0133 & 0.0632 & 0.0646\\
 & TMLE & 0.1363 & 0.1360 & 0.0003 & 0.0616 & 0.0616\\
\midrule
\multirow{4}{*}{$Q$ correct; $g$ misspecified} & MLE  & 0.1363 & 0.1367 & 0.0004 & 0.0393 & 0.0393\\
 & IPW  & 0.1363 & 0.1385 & 0.0023 & 0.0606 & 0.0606\\
 & AIPW & 0.1363 & 0.1360 & 0.0003 & 0.0472 & 0.0472\\
 & TMLE & 0.1363 & 0.1357 & 0.0005 & 0.0482 & 0.0482\\
\midrule
\multirow{4}{*}{Both misspecified} & MLE  & 0.1363 & 0.2363 & 0.1000 & 0.0609 & 0.1171\\
 & IPW  & 0.1363 & 0.1385 & 0.0023 & 0.0606 & 0.0606\\
 & AIPW & 0.1363 & 0.1321 & 0.0041 & 0.0612 & 0.0613\\
 & TMLE & 0.1363 & 0.1246 & 0.0117 & 0.0584 & 0.0595\\
\midrule
\multirow{4}{*}{Both correctly specified} & MLE  & 0.1363 & 0.1367 & 0.0004 & 0.0393 & 0.0393\\
 & IPW  & 0.1363 & 0.1350 & 0.0012 & 0.0593 & 0.0593\\
 & AIPW & 0.1363 & 0.1360 & 0.0003 & 0.0488 & 0.0488\\
 & TMLE & 0.1363 & 0.1362 & 0.0001 & 0.0508 & 0.0508\\
\bottomrule
\multicolumn{3}{l}{\footnotesize{$^*$ ``correct'' means ``correctly specified''}}
\end{tabular}
\end{table}

\begin{table}[htbp]
\centering
\caption{TMLE confidence-interval performance for a given $h$}
\label{tab:four-scenario-inference}
\begin{tabular}{llrr}
\toprule
Scenario & Interval method & Mean SE & Coverage\\
\midrule
\multirow{3}{*}{$Q$ misspecified; $g$ correct} & EIF-Wald & 0.057 & 0.910\\
 & TB-Wald & 0.069 & 0.958\\
 & TB-percentile & --- & 0.968\\
\midrule
\multirow{3}{*}{$Q$ correct; $g$ misspecified} & EIF-Wald & 0.044 & 0.915\\
 & TB-Wald & 0.048 & 0.937\\
 & TB-percentile & --- & 0.939\\
\midrule
\multirow{3}{*}{Both misspecified} & EIF-Wald & 0.056 & 0.893\\
 & TB-Wald & 0.062 & 0.922\\
 & TB-percentile & --- & 0.944\\
\midrule
\multirow{3}{*}{Both correctly specified} & EIF-Wald & 0.046 & 0.902\\
 & TB-Wald & 0.052 & 0.943\\
 & TB-percentile & --- & 0.950\\
\bottomrule
\end{tabular}
\end{table}

As shown in Table \ref{tab:four-scenario-point}, when $Q$ was misspecified but $g$ was correctly specified, TMLE had the smallest bias and a relatively small RMSE among the four estimators. This illustrates the finite-sample stabilizing effect of the targeting step under outcome-model misspecification. When $Q$ was correctly specified but $g$ was misspecified, MLE had the smallest RMSE, and AIPW and TMLE were close. When both nuisance models were misspecified, neither AIPW nor TMLE was protected from bias, although TMLE had slightly smaller RMSE than AIPW in this simulation. When both nuisance models were correctly specified, all doubly robust estimators had very small bias, and AIPW and TMLE had similar RMSE.

As shown in Table \ref{tab:four-scenario-inference}, the EIF-Wald interval exhibited some undercoverage across all four scenarios. This finite-sample limitation has been discussed previously \citep{van2018targeted}, where the target bootstrap was proposed as a practical alternative. In our implementation of the target bootstrap, the initial nuisance estimates were held fixed, and only the TMLE fluctuation step was refitted within each stratified bootstrap sample. Because the percentile interval is constructed directly from the empirical quantiles of the bootstrap distribution, it does not have a single associated standard-error estimate. Both target bootstrap methods improved coverage relative to the EIF-Wald interval across all four scenarios.

\section{Discussion}

This paper makes three primary contributions:
\begin{enumerate}
\item We review the principal data sources used to construct ECAs in externally controlled trials.
\item We develop TMLEs for a general class of balancing-weight estimands $\tau_h$, corresponding to prespecified tilting functions $h(x)$, within the targeted learning framework.
\item We demonstrate that the proposed framework extends naturally beyond ECTs to two additional settings: RCTs augmented with external covariate information defining a target population, and ITCs that combine individual patient data with aggregate data.
\end{enumerate}

The proposed TMLEs substantially broaden the scope of targeted learning by accommodating a general class of balancing-weight estimands defined through prespecified tilting functions. As a result, the proposed framework provides a unified and flexible approach to estimating treatment effects in target populations commonly encountered in externally controlled trials and related comparative effectiveness studies.

We distinguish between two scenarios. In the first, the titling function is treated as fixed when defining the population attribute of the target estimand. In the second, the titling function is treated as unknown, and the resulting estimated population is viewed as an estimate of an underlying true target population. This paper focuses on the first scenario; the second is considered in \cite{fang2026targeted}.

Like other balancing-weight methods, the proposed estimators rely on standard causal identification assumptions, including consistency, conditional exchangeability, and positivity. In practice, limited covariate overlap may lead to unstable weights and reduced estimation efficiency, while unmeasured confounding or effect modification may introduce bias despite the use of doubly robust estimation.

Several extensions merit future investigation. These include developing diagnostic tools for assessing covariate overlap and identification assumptions, extending the methodology to longitudinal, survival, clustered, and multiple-treatment settings, and incorporating more flexible machine learning algorithms through the Super Learner to improve estimation in high-dimensional applications while retaining the desirable theoretical properties of TMLE.

\section{Disclosure}

The initial draft of the \textsf{R} code used to generate the simulation results presented in this section was produced with the assistance of ChatGPT 5.2; the code was subsequently revised by the authors.

\bibliographystyle{apalike}
\bibliography{references}

@article{birnbaum1962foundations,
  title={On the foundations of statistical inference},
  author={Birnbaum, Allan},
  journal={Journal of the American Statistical Association},
  volume={57},
  number={298},
  pages={269--306},
  year={1962},
  publisher={Taylor \& Francis}
}

@article{fang2026targeted,
  title={Targeted Learning for Tilted Average Treatment Effects},
  author={Fang, Yixin},
  journal={Statistics in Biopharmaceutical Research},
  volume={Published},
  pages={1--48},
  year={2026},
  publisher={Taylor \& Francis}
}

@article{hirano2003efficient,
  title={Efficient estimation of average treatment effects using the estimated propensity score},
  author={Hirano, Keisuke and Imbens, Guido W and Ridder, Geert},
  journal={Econometrica},
  volume={71},
  number={4},
  pages={1161--1189},
  year={2003},
  publisher={Wiley Online Library}
}

@article{zhou2020propensity,
  title={Propensity score weighting under limited overlap and model misspecification},
  author={Zhou, Yunji and Matsouaka, Roland A and Thomas, Laine},
  journal={Statistical methods in medical research},
  volume={29},
  number={12},
  pages={3721--3756},
  year={2020},
  publisher={SAGE Publications Sage UK: London, England}
}

@article{liliang2013weighting,
  title={A weighting analogue to pair matching in propensity score analysis},
  author={Li, Liang and Greene, Tom},
  journal={The international journal of biostatistics},
  volume={9},
  number={2},
  pages={215--234},
  year={2013},
  publisher={De Gruyter}
}

@article{rytgaard2024one,
  title={One-step targeted maximum likelihood estimation for targeting cause-specific absolute risks and survival curves},
  author={Rytgaard, Helene CW and van der Laan, Mark J},
  journal={Biometrika},
  volume={111},
  number={1},
  pages={129--145},
  year={2024},
  publisher={Oxford University Press}
}

@article{cheng2022addressing,
  title={Addressing extreme propensity scores in estimating counterfactual survival functions via the overlap weights},
  author={Cheng, Chao and Li, Fan and Thomas, Laine E and Li, Fan},
  journal={American journal of epidemiology},
  volume={191},
  number={6},
  pages={1140--1151},
  year={2022},
  publisher={Oxford University Press}
}

@article{liu2025leveraging,
  title={Leveraging External Controls in Clinical Trials: Estimands, Estimation, Assumptions},
  author={Liu, Bo and Thomas, Laine and Holman, Rury R and Li, Fan},
  journal={arXiv preprint arXiv:2503.21081},
  year={2025}
}

@article{cao2025using,
  title={Using overlap weights to address extreme propensity scores in estimating restricted mean counterfactual survival times},
  author={Cao, Zhiqiang and Ghazi, Lama and Mastrogiacomo, Claudia and Forastiere, Laura and Wilson, F Perry and Li, Fan},
  journal={American Journal of Epidemiology},
  volume={194},
  number={8},
  pages={2402--2411},
  year={2025},
  publisher={Oxford University Press}
}

@article{liu2024average,
  title={Average treatment effect on the treated, under lack of positivity},
  author={Liu, Yi and Li, Huiyue and Zhou, Yunji and Matsouaka, Roland A},
  journal={Statistical Methods in Medical Research},
  volume={33},
  number={10},
  pages={1689--1717},
  year={2024},
  publisher={SAGE Publications Sage UK: London, England}
}

@article{wang2025rate,
  title={Rate doubly robust estimation for weighted average treatment effects},
  author={Wang, Yiming and Liu, Yi and Yang, Shu},
  journal={Journal of Causal Inference},
  volume={13},
  number={1},
  pages={20240073},
  year={2025},
  publisher={De Gruyter}
}

@book{van2000asymptotic,
  title={Asymptotic statistics},
  author={Van der Vaart, Aad W},
  volume={3},
  year={2000},
  publisher={Cambridge university press}
}

@article{li2018balancing,
  title={Balancing covariates via propensity score weighting},
  author={Li, Fan and Morgan, Kari Lock and Zaslavsky, Alan M},
  journal={Journal of the American Statistical Association},
  volume={113},
  number={521},
  pages={390--400},
  year={2018},
  publisher={Taylor \& Francis}
}

@article{gruber2012tmle,
  title={tmle: An {R} package for targeted maximum likelihood estimation},
  author={Gruber, Susan and van der Laan, Mark},
  journal={Journal of Statistical Software},
  volume={51},
  pages={1--35},
  year={2012}
}

@article{andersen2004regression,
  title={Regression analysis of restricted mean survival time based on pseudo-observations},
  author={Andersen, Per Kragh and Hansen, Mette Gerster and Klein, John P},
  journal={Lifetime Data Analysis},
  volume={10},
  pages={335--350},
  year={2004},
  publisher={Springer}
}

@book{fang2024causal,
  title={Causal Inference in Pharmaceutical Statistics},
  author={Fang, Yixin},
  year={2024},
  publisher={CRC Press}
}

@book{fang2026every,
  title={Every Coin Has Two Sides: An Introduction to Causal Inference in Pharmaceutical Statistics},
  author={Fang, Yixin},
  year={2026},
  publisher={CRC Press}
}

@article{phillippo2016nice,
  title={NICE DSU technical support document 18: methods for population-adjusted indirect comparisons in submissions to NICE},
  author={Phillippo, David and Ades, A E and Dias, Sofia and Palmer, Stephen and Abrams, Keith R and Welton, Nicky J},
  year={2016},
  journal={NICE Decision Support Unit}
}

@article{signorovitch2010comparative,
  title={Comparative effectiveness without head-to-head trials: a method for matching-adjusted indirect comparisons applied to psoriasis treatment with adalimumab or etanercept},
  author={Signorovitch, James E and Wu, Eric Q and Yu, Andrew P and Gerrits, Charles M and Kantor, Evan and Bao, Yanjun and Gupta, Shiraz R and Mulani, Parvez M},
  journal={Pharmacoeconomics},
  volume={28},
  pages={935--945},
  year={2010},
  publisher={Springer}
}

@article{ich2020e9,
  title={{ICH E9 (R1)} addendum on estimands and sensitivity analysis in clinical trials to the Guideline on statistical principles for clinical trials},
  author={ICH},
  journal={Accessed: 17th February},
  year={2020}
}

@book{imbens2015causal,
  title={Causal inference in statistics, social, and biomedical sciences},
  author={Imbens, Guido W and Rubin, Donald B},
  year={2015},
  publisher={Cambridge University Press}
}

@book{hernan2020causal,
  title={Causal Inference: What If},
  author={Hern\'an, Miguel A and Robins, James M},
  year={2020},
  publisher={Boca Raton: Chapman \& Hall/CRC.}
}

@book{van2011targeted,
  title={Targeted Learning: Causal Inference for Observational and Experimental Data},
  author={van der Laan, Mark J and Rose, Sherri},
  year={2011},
  publisher={Springer Science \& Business Media}
}

@article{rosenbaum1983central,
  title={The central role of the propensity score in observational studies for causal effects},
  author={Rosenbaum, Paul R and Rubin, Donald B},
  journal={Biometrika},
  volume={70},
  number={1},
  pages={41--55},
  year={1983},
  publisher={Oxford University Press}
}

@article{hernan2010hazards,
  title={The hazards of hazard ratios},
  author={Hern{\'a}n, Miguel A},
  journal={Epidemiology (Cambridge, Mass.)},
  volume={21},
  number={1},
  pages={13},
  year={2010},
  publisher={NIH Public Access}
}

@article{li2013weighting,
  title={On weighting approaches for missing data},
  author={Li, Lingling and Shen, Changyu and Li, Xiaochun and Robins, James M},
  journal={Statistical Methods in Medical Research},
  volume={22},
  number={1},
  pages={14--30},
  year={2013}
}

@book{bickel1993efficient,
  title={Efficient and Adaptive Estimation for Semiparametric Models},
  author={Bickel, Peter J and Klaassen, Chris AJ and Bickel, Peter J and Ritov, Ya’acov and Klaassen, J and Wellner, Jon A and Ritov, YA'Acov},
  year={1993},
  publisher={Springer}
}

@article{fang2020two,
  title={Two basic statistical strategies of conducting causal inference in real-world studies},
  author={Fang, Yixin},
  journal={Contemporary Clinical Trials},
  volume={99},
  pages={106193},
  year={2020},
  publisher={Elsevier}
}

@misc{FDA_ECT_2023,
  author    = {{FDA}},
  title     = {Considerations for the Design and Conduct of Externally Controlled Trials for Drug and Biological Products Guidance for Industry},
  howpublished = {\url{https://www.fda.gov/media/164960/download}},
  year      = {2023},
}

@book{tsiatis2006semiparametric,
  title={Semiparametric Theory and Missing Data},
  author={Tsiatis, A},
  year={2006},
  publisher={Springer}
}

@article{van2007super,
  title={Super learner},
  author={van der Laan, Mark J and Polley, Eric C and Hubbard, Alan E},
  year={2007},
  journal={Statistical Applications in Genetics and Molecular Biology},
  volume={6},
  number={1},
  publisher={bepress}
}

@book{van2018targeted,
  title={Targeted Learning in Data Science},
  author={van der Laan, Mark J and Rose, Sherri},
  year={2018},
  publisher={Springer}
}

@article{phillips2023practical,
  title={Practical considerations for specifying a super learner},
  author={Phillips, Rachael V and van der Laan, Mark J and Lee, Hana and Gruber, Susan},
  journal={International Journal of Epidemiology},
  volume={52},
  number={4},
  pages={1276--1285},
  year={2023},
  publisher={Oxford University Press}
}

@article{colnet2024causal,
  title={Causal inference methods for combining randomized trials and observational studies: a review},
  author={Colnet, B{\'e}n{\'e}dicte and Mayer, Imke and Chen, Guanhua and Dieng, Awa and Li, Ruohong and Varoquaux, Ga{\"e}l and Vert, Jean-Philippe and Josse, Julie and Yang, Shu},
  journal={Statistical Science},
  volume={39},
  number={1},
  pages={165--191},
  year={2024},
  publisher={Institute of Mathematical Statistics}
}

@article{cole2010generalizing,
  title={Generalizing evidence from randomized clinical trials to target populations: the ACTG 320 trial},
  author={Cole, Stephen R and Stuart, Elizabeth A},
  journal={American Journal of Epidemiology},
  volume={172},
  number={1},
  pages={107--115},
  year={2010},
  publisher={Oxford University Press}
}

@article{dahabreh2021study,
  title={Study designs for extending causal inferences from a randomized trial to a target population},
  author={Dahabreh, Issa J and Haneuse, Sebastien JP A and Robins, James M and Robertson, Sarah E and Buchanan, Ashley L and Stuart, Elizabeth A and Hern{\'a}n, Miguel A},
  journal={American Journal of Epidemiology},
  volume={190},
  number={8},
  pages={1632--1642},
  year={2021},
  publisher={Oxford University Press}
}

@article{lee2023improving,
  title={Improving trial generalizability using observational studies},
  author={Lee, Dasom and Yang, Shu and Dong, Lin and Wang, Xiaofei and Zeng, Donglin and Cai, Jianwen},
  journal={Biometrics},
  volume={79},
  number={2},
  pages={1213--1225},
  year={2023},
  publisher={Oxford University Press}
}

@article{colnet2022causal,
  title={Causal effect on a target population: A sensitivity analysis to handle missing covariates},
  author={Colnet, B{\'e}n{\'e}dicte and Josse, Julie and Varoquaux, Ga{\"e}l and Scornet, Erwan},
  journal={Journal of Causal Inference},
  volume={10},
  number={1},
  pages={372--414},
  year={2022},
  publisher={De Gruyter}
}

@article{karrison1987restricted,
  title={Restricted mean life with adjustment for covariates},
  author={Karrison, Theodore},
  journal={Journal of the American Statistical Association},
  volume={82},
  number={400},
  pages={1169--1176},
  year={1987},
  publisher={Taylor \& Francis}
}

@article{jin2026framework,
  title={A Framework for Estimating Restricted Mean Survival Time Difference using Pseudo-observations},
  author={Jin, Man and Fang, Yixin},
  journal={arXiv preprint arXiv:2601.06296},
  year={2026}
}

@book{he2023real,
  title={Real-World Evidence in Medical Product Development},
  author={He, Weili and Fang, Yixin and Wang, Hongwei},
  year={2023},
  publisher={Springer}
}

@article{yang2021subgroup,
  author    = {Yang, Shu and Lorenzi, Elisa and Papadogeorgou, Georgia and Li, Fan},
  title     = {Propensity score weighting for causal subgroup analysis},
  journal   = {Statistics in Medicine},
  year      = {2021},
  volume    = {40},
  number    = {20},
  pages     = {4294--4309},
  doi       = {10.1002/sim.9029}
}

@article{zeng2023pseudo,
  author    = {Zeng, Shuxi and Li, Fan and Hu, Liang and others},
  title     = {Propensity score weighting analysis of survival outcomes using pseudo-observations},
  journal   = {Statistica Sinica},
  year      = {2023},
  volume    = {33},
  number    = {3},
  pages     = {2161--2184},
  doi       = {10.5705/ss.202021.0175}
}

@article{zhou2022psweight,
  author    = {Zhou, Tianying and Tong, Guanhua and Li, Fan and Thomas, L. Elizabeth and Li, Fan},
  title     = {PSweight: An R Package for Propensity Score Weighting Analysis},
  journal   = {The R Journal},
  year      = {2022},
  volume    = {14},
  number    = {1},
  pages     = {282--300},
  doi       = {10.32614/RJ-2022-011}
}

@article{cao2024overlap,
  author    = {Cao, Zhe and Ghazi, Leila and Forastiere, Laura and Wilson, Francis P. and Li, Fan},
  title     = {Using overlap weights to address extreme propensity scores in estimating restricted mean counterfactual survival times},
  journal   = {American Journal of Epidemiology},
  year      = {2024},
  volume    = {194},
  number    = {8},
  pages     = {2402--2411},
  doi       = {10.1093/aje/kwae416}
}

@article{liu2025positivity,
  author    = {Liu, Yifan and Wang, Ying and Gao, Yu and others},
  title     = {A tutorial for propensity score weighting methods under violations of the positivity assumption},
  journal   = {Statistics in Medicine},
  year      = {2025},
  note      = {Online ahead of print},
  doi       = {10.1002/sim.70329}
}

\newpage

\section*{Appendix}
\setcounter{table}{0}
\renewcommand{\thetable}{A\arabic{table}}

\setcounter{figure}{0}
\renewcommand{\thefigure}{A\arabic{figure}}

\subsection*{A.1 Efficient Influence Function}

Define
\[
\tau_h=\frac{\theta_h}{\eta_h}, \qquad
\theta_h=\mathbb{E}\{h(X)[Q(X,1)-Q(X,0)]\}, \qquad
\eta_h=\mathbb{E}[h(X)].
\]
The EIFs for $\theta_h$ and $\eta_h$ are
\begin{align*}
\phi_1(X,A,Y)
&=
h(X)[Q(X,1)-Q(X,0)]-\theta_h \\
&\quad +
\left[
\frac{h(X)A}{g(X)}
-
\frac{h(X)(1-A)}{1-g(X)}
\right][Y-Q(X,A)],
\\
\phi_2(X)&=h(X)-\eta_h.
\end{align*}
By the delta method for pathwise differentiable functionals,
\[
\phi_h(X,A,Y)
=
\frac{1}{\eta_h}\big[\phi_1(X,A,Y)-\tau_h\phi_2(X)\big].
\]
After simplification, the EIF for $\tau_h$ is
\begin{align*}
\phi_h(X,A,Y)
&=
h^*(X)[Q(X,1)-Q(X,0)-\tau_h] \\
&\quad +
\left[
\frac{h^*(X)A}{g(X)}
-
\frac{h^*(X)(1-A)}{1-g(X)}
\right][Y-Q(X,A)].
\end{align*}

\newpage

\subsection*{A.2 Partial \textsf{R} Code}

\begin{lstlisting}
## Some simple functions
expit  <- function(x) 1 / (1 + exp(-x))
logit  <- function(p) log(p / (1 - p))

## Base clever covariate
HAW <- A / g1W - (1 - A) / g0W 

## For a given function h(W), h = (h(W1), ..., h(Wn))
one_target <- function(h) { 
  h_star <- h / mean(h)                # normalize h 
  HAW_star <- h_star * HAW             # new clever covariate H_star 
   
  logitQAW <- logit(QAW)               
  fluc <- glm(Y ~ -1 + offset(logitQAW) + Htilde, family = binomial())
  eps  <- as.numeric(coef(fluc)[1])    # epsilon hat is obtained

  QstarAW <- expit(logitQAW + eps * HAW_star)
  Qstar1W <- expit(logit(Q1W) + eps * (h_star / g1W))
  Qstar0W <- expit(logit(Q0W) - eps * (h_star / g0W))
}

tauW <- Qstar1W - Qstar0W

## Plug-in estimator using h_star weights
  tau_hat <- sum(h_star * tauW)

## Influence curve and standard error  
  IC <- h_star * (tauW + HAW * (Y - QstarAW) - tau_hat)
  se <- sqrt(var(IC) / n)
\end{lstlisting}

\end{document}